\documentclass[letterpaper, 10 pt, conference]{ieeeconf}  

\IEEEoverridecommandlockouts                           
\title{\LARGE \bf
ConPoSe: LLM-Guided Contact Point Selection for Scalable \\ Cooperative Object Pushing 
}

\author{Noah Steinkrüger$^{1}$, Nisarga Nilavadi$^{1}$, Wolfram Burgard$^{1}$, and Tanja Katharina Kaiser$^{1}$% 
\thanks{$^{1}$All authors are with the Department of Computer Science and Artificial Intelligence, 
        University of Technology Nuremberg, Germany. 
        {\tt\small \textit{{firstname}}.\textit{{lastname}}@utn.de}} %
}

\usepackage{color, colortbl}
\usepackage[table, x11names]{xcolor}
\usepackage{amsfonts}
\usepackage{amsmath}
\DeclareMathOperator{\arctantwo}{arctan2}
\usepackage{graphicx}
\usepackage{siunitx}
\usepackage{stackengine}
\usepackage{multirow}
\usepackage{adjustbox}    

\usepackage{listings}
\usepackage{comment}
\usepackage{algorithm}
\usepackage{algpseudocode}
\algrenewcommand\algorithmicrequire{\textbf{Input:}}
\algrenewcommand\algorithmicensure{\textbf{Output:}}
\usepackage{booktabs}
\usepackage{tikz}
\usepackage{pifont}
\newcommand{\etal}{~\emph{et al.}}
\usepackage[caption=false]{subfig}

\newcommand{\cp}{c}
\newcommand{\robots}{N}

\usepackage{pgfplots}
\usepackage{pgfplotstable}
\usetikzlibrary{patterns}
\pgfplotsset{compat=1.18}

\begin{document}

\maketitle
\thispagestyle{empty}
\pagestyle{empty}

\def\mathdefault#1{#1} 

%%%%%%%%%%%%%%%%%%%%%%%%%%%%%%%%%%%%%%%%%%%%%%%%%%%%%%%%%%%%%%%%%%%%%%%%%%%%%%%%
\begin{abstract}
Object transportation in cluttered environments is a fundamental task in various domains, including domestic service and warehouse logistics. 
In cooperative object transport, multiple robots must coordinate to move objects that are too large for a single robot. 
One transport strategy is pushing, which only requires simple robots.
However, careful selection of robot-object contact points is necessary to push the object along a preplanned path. 
Although this selection can be solved analytically, the solution space grows combinatorially with the number of robots and object size, limiting scalability. 
Inspired by how humans rely on common-sense reasoning for cooperative transport, we propose combining the reasoning capabilities of Large Language Models with local search to select suitable contact points. 
Our LLM-guided local search method for \underline{con}tact \underline{po}int \underline{se}lection, ConPoSe, successfully selects contact points for a variety of shapes, including cuboids, cylinders, and T-shapes. 
We demonstrate that ConPoSe scales better with the number of robots and object size than the analytical approach, and also outperforms pure LLM-based selection. 
\end{abstract}
%%%%%%%%%%%%%%%%%%%%%%%%%%%%%%%%%%%%%%%%%%%%%%%%%%%%%%%%%%%%%%%%%%%%%%%%%%%%%%%%

\section{INTRODUCTION}

Object transportation is a fundamental task in many robotic applications. 
Examples are construction, delivery, manufacturing, and search and rescue. 
For large, heavy, or bulky objects, the cooperative transportation by multiple robots may be required~\cite{tuci2018}.
In pushing-only approaches to object transportation, simple robots without means of grasping are sufficient. 
However, those strategies require close coordination among the robots to achieve efficient object transport. 
Robots may lose contact with the object, and frictional, gravitational, and dynamical forces affect the transportation direction. 
Thus, a main challenge is selecting and re-adjusting robot-object contact points to push the object as quickly as possible along a preplanned path toward its goal position. 
Existing works address these challenges by proposing novel algorithms~\cite{NEDJAH2025126610}, learning-based approaches~\cite{feng2024learningmultiagentlocomanipulationlonghorizon, Gross2004}, and hybrid optimization techniques~\cite{tang2024collaborativeplanarpushingpolytopic}.
However, research on object transportation scenarios often focuses on simplified conditions, for example, small-scale Multi-Robot Systems (MRSs)~\cite{feng2024learningmultiagentlocomanipulationlonghorizon, song2025collabotvisionlanguageguidedsimultaneous}, a single object shape~\cite{NEDJAH2025126610}, or environments with simplified obstacles~\cite{NEDJAH2025126610, Gross2004, Bertoncelli2021}.
Other approaches presuppose detailed prior knowledge of the pushed object’s intrinsic properties~\cite{tang2024collaborativeplanarpushingpolytopic}, or make use of holonomic robots, which facilitates the maintenance of stable robot–object contact points~\cite{feng2024learningmultiagentlocomanipulationlonghorizon, tang2024collaborativeplanarpushingpolytopic}.

By contrast, humans demonstrate the ability to cooperatively transport a wide range of objects, relying on limited object information and common-sense reasoning.
To endow robots with similar reasoning and generalization capabilities, pre-trained foundation models, such as large language models (LLMs) and vision language models (VLMs), are increasingly being adopted~\cite{Firoozi2024}. 
Applications of foundation models in MRSs include task planning~\cite{kannan2024smartllmsmartmultiagentrobot}, visual semantic navigation~\cite{shen2025enhancingmultirobotsemanticnavigation}, pattern formation~\cite{venkatesh2024zerocapzeroshotmultirobotcontext}, and collaborative manipulation by grasping~\cite{song2025collabotvisionlanguageguidedsimultaneous}.
The latter two application domains involve complex geometrical reasoning, which is also needed for cooperative object transportation.
We show in this paper that the common-sense reasoning of foundation models can also be used for solving the cooperative object pushing task.

 \begin{figure}
     \centering
     \includegraphics[width=\linewidth]{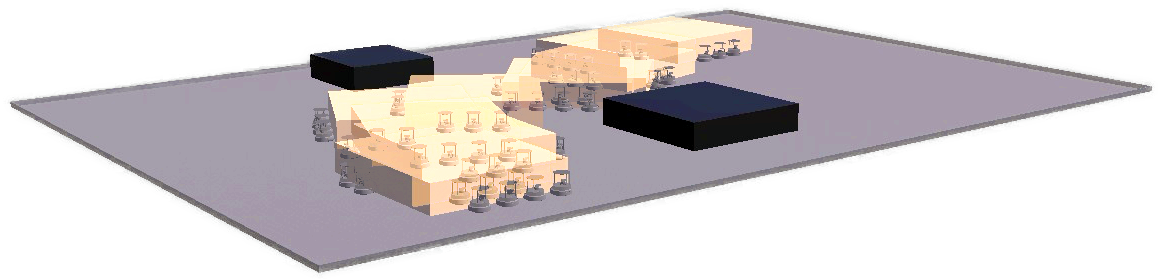}
     \caption{Visualization of the object pushing task: a cuboid is pushed from its initial position in the front-left to the goal in the back-right corner.}
     \label{fig:visualization_scenario}
 \end{figure}

In this paper, we propose ConPoSe, an LLM-guided contact point selection method for cooperative object pushing. 
In this method, we combine the common-sense reasoning capabilities of LLMs with local search to select robot-object contact points for pushing an object along a preplanned path (see Fig.~\ref{fig:visualization_scenario}). 
The required prior knowledge about the object is limited to its shape and center of mass; all remaining information is directly inferred using the environment state. 
Our experimental results demonstrate that ConPoSe achieves better time scalability compared to analytically computing the contact points, and still reaches near-perfect success rates.  
In comparison to pure LLM-based contact point selection, ConPoSe requires fewer contact point switches to follow the preplanned path, making it more efficient than both baselines. 
In summary, our contributions are as follows: 

\begin{enumerate}
    \item We propose ConPoSe, a contact point selection method for cooperative object pushing that combines the common-sense reasoning capabilities of foundation models with local search. 
    \item We present evaluations in a realistic simulator with up to 15~robots, three different object shapes, and large-scale scenarios with object path lengths between \SI{10}{m} and \SI{100}{m}. 
    \item We show that ConPoSe exhibits superior time scalability compared to the analytical baseline, while still achieving competitive success rates.
    In addition, we show that ConPoSe is more efficient in selecting contact points than a pure LLM-based selection method. 
    \item We will release our code. 
\end{enumerate}

\section{RELATED WORK}

\subsection{Cooperative Object Pushing}

Approaches to cooperative object transport can be categorized into three different transportation strategies: pushing, grasping, and caging~\cite{tuci2018}. 
In this work, we focus on object pushing with multiple simple autonomous ground robots.
Existing approaches range from analytical to learning-based methods, however, many rely on simplifying assumptions. 
For example, the algorithm proposed by Nedjah\etal~\cite{NEDJAH2025126610} is limited to pushing cylindrical objects and does not generalize to other shapes. 
Like Nedjah\etal~\cite{NEDJAH2025126610}, Gro{\ss} and Dorigo~\cite{Gross2004} and Bertoncelli and Sabattini~\cite{Bertoncelli2021} only consider environments with simplified obstacles. 
Gro{\ss} and Dorigo~\cite{Gross2004} evolve artificial neural networks for the transport of cuboid and cylindrical objects, while Bertoncelli and Sabattini~\cite{Bertoncelli2021} propose a Voronoi-based coverage control method applicable to different object shapes. 
By contrast, Tang\etal~\cite{tang2024collaborativeplanarpushingpolytopic} introduce a hierarchical hybrid search framework for pushing polytopic and non-polytopic objects in cluttered environments. 
However, their method requires detailed prior knowledge of the object properties, such as friction coefficients. 
As they highlight, pushing forces and friction lack an analytical form for multiple bodies under contact, introducing model uncertainties. 
Consequently, online adaptation remains essential for pushing the object along a preplanned path. 
Humans, in comparison, can transport objects using limited prior information about the object, by leveraging world knowledge and common-sense reasoning. 
Motivated by this, we propose to exploit the generalization and common-sense reasoning capabilities of foundation models and combine it with the strengths of local search for solving the computationally hard problem of selection suitable contact points for pushing an object along a preplanned path.  
In addition, we employ round differential drive robots, similar to Gro{\ss} and Dorigo~\cite{Gross2004}, which makes maintaining stable contact points more challenging~\cite{RAL_Tang_2023} than in approaches that rely on holonomic robots~\cite{feng2024learningmultiagentlocomanipulationlonghorizon, tang2024collaborativeplanarpushingpolytopic}.

\subsection{Foundation Models in Robotics}

Foundation models have demonstrated potential to enhance adaptability and generalization in robotics and multi-robot systems~\cite{Firoozi2024}.
Their spatial reasoning capabilities have been leveraged in robot navigation, for example through end-to-end navigation policies~\cite{goetting2024endtoend}. 
Visual annotations in image input to VLMs can further improve zero-shot spatial reasoning in tasks such as navigation, grasping, and object rearrangement~\cite{NasirianyX0XL0X24}.  
Foundation models have also been applied for geometrical reasoning tasks in both single and multi-robot systems. 
Examples include generating grasp poses for task-oriented manipulation~\cite{tang2025} and collaborative manipulation~\cite{song2025collabotvisionlanguageguidedsimultaneous}.
In multi-robot systems, foundation models are used for a variety of applications, such as task planning~\cite{kannan2024smartllmsmartmultiagentrobot, zhang2024building}, visual semantic navigation~\cite{shen2025enhancingmultirobotsemanticnavigation}, and perception~\cite{blumenkamp2024covisnet}. 
Most closely related to our work is the foundation model-based zero-shot approach to pattern formation related to an obstacle proposed by Venkatesh and Min~\cite{venkatesh2024zerocapzeroshotmultirobotcontext}.
In their work, an LLM determines the (x,y)-coordinates for each robot based on a shape description and a pattern formation prompt generated by a VLM. 
Unlike pattern formation, collective pushing also requires consideration of the global pushing direction when selection robot-object contact points. 
Moreover, robots may need to dynamically switch contact points multiple times during task execution. 
To the best of our knowledge, we are the first to propose a foundation model-based approach to collective pushing.

\begin{figure*}[t]
    \centering
    \includegraphics[width=\linewidth]{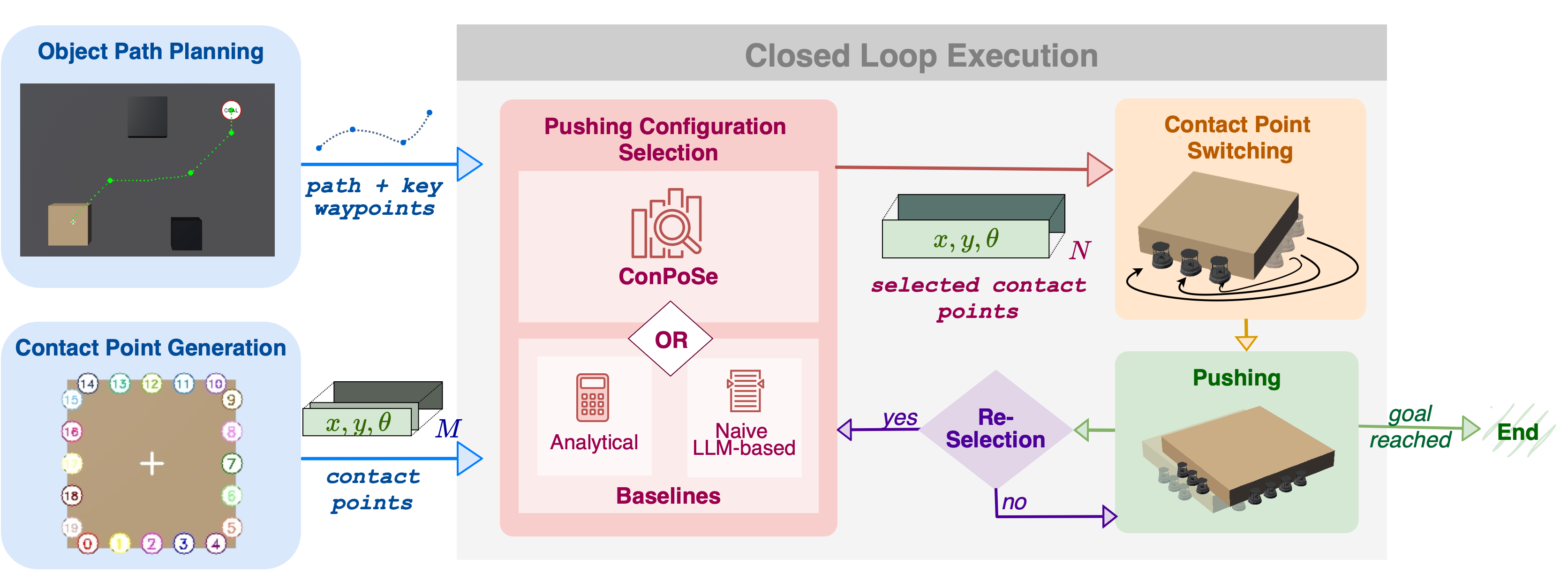} 
    \caption{Overview of our approach. At the start of each experiment, we generate a global object path and key waypoints (Sec.~\ref{sec:obj_path_planning}), along with a set of \(M\)~candidate contact points on the object contour (Sec.~\ref{sec:cp_generation}). Object pushing then proceeds in closed loop. Contact points are selected using our proposed LLM-guided local search approach, ConPoSe, or using one of our two baselines: naive LLM-based selection or analytical computation of the contact points (Sec.~\ref{sec:push_config_selection}), after which the robots switch to their assigned contact points (Sec.~\ref{sec:cp_switching}). The object is then pushed (Sec.~\ref{sec:pushing}) until either the goal is reached or re-selection becomes necessary (e.g., due to deviations from the planned path; Sec.~\ref{sec:online_adaptation}).}     
    \label{fig:Overview}
\end{figure*}

\section{PROBLEM FORMULATION}

We address the problem of pushing an object~\(O\) along a preplanned path ~\(S_O = \{s_O^1, \dots, s_O^{J}\}\) of \(J\)~waypoints~\(s_O^j = (x_O^j, y_O^j)\) from its initial position~\(s_O^1\) to a goal position~\(s_O^{J}\) in a 2D~environment~\(\mathcal{W} \subset \mathbb{R}^2\), using a homogeneous MRS of \(\robots\)~round differential drive robots.
Round robots interact with objects via single-point contacts, making it challenging for nonholonomic round robots to maintain stable contact points with the object~\cite{RAL_Tang_2023}. 
The environment contains static obstacles with known shapes and positions. 
Collisions with obstacles must be avoided. 

The object state at time~\(t \geq 0\) is given by the pose
\begin{equation}
p_O(t) = (\mathbf{x}_O(t), \theta_O(t)),\, \mathbf{x}_O(t) = (x_O(t), y_O(t)), 
\end{equation}
where \(\theta_O(t)\) is the object rotation. 
The object shape is described by its footprint~\(\mathcal{B}\).
Its physical properties, including its mass, are unknown.
We assume a uniform mass distribution, so that the geometrical object center and the center of mass overlap. 
Each robot~\(n\) interacts with the object by applying a pushing force at a contact point~\(c^n \in \delta \mathcal{B}\) along its boundary. 
All robots exert pushing forces of equal magnitude, \(|\mathbf{F}^n_\text{Bot}| = |\mathbf{F}_\text{Bot}|\), directed along the inward normal to~\(\delta \mathcal{B}\). 
The contact point of robot~\(n\) is represented by 
\begin{align}
    c^n &= (\mathbf{x}^n_\text{CP}, \theta_\text{CP}^n), \, \mathbf{x}^n_\text{CP} = (x_\text{CP}^n, y_\text{CP}^n)
\end{align}
where \(\theta_\text{CP}^n\) denotes the direction in which the pushing force is applied. 
Uncertainties regarding robots maintaining their contact points can lead to unexpected and unknown deviations in the actual applied pushing forces. 

Each robot pushes the object at a different contact point~(CP), resulting in a \textit{pushing configuration} 
\begin{align}
    \mathcal{P} &= \{\cp^{1}, \dots, \cp^{N}\}  
\end{align}
of \(N\)~unique contact points~\(c^n, \, n\in N\). 
Since contact points and pushing configurations are not hand-picked, the number of possible pushing configurations grows combinatorially with the size of both the object and the MRS. 

The state of each robot~\(n\) is defined by its pose 
\begin{align}
   p^n_{\text{Bot}}(t) &= (\mathbf{x}^n_{\text{Bot}}(t), \theta^n_{\text{Bot}}(t))\, , \\
   \mathbf{x}^n_{\text{Bot}}(t) &= (x^n_{\text{Bot}}(t), y^n_{\text{Bot}}(t))  
\end{align}

with heading~\(\theta^n_{\text{Bot}}(t)\).
Each robot~\(n\) is controlled via a linear velocity~\(v^n(t)\) along the x-axis and an angular velocity~\(\omega^n(t)\) around the z-axis, and has an assigned contact point~\(\hat{\cp}^n\). 

Our objective is to determine a sequence of \(Z\)~pushing configurations~\(\mathcal{P}^z \in \{\mathcal{P}^1, \dots, \mathcal{P}^Z\}\) that moves the object along the preplanned path~\(S_O\). 
The number of pushing configurations~\(Z\) should be as small as possible to reduce time-consuming contact point switches, while ensuring collision-free object transport by close adherence to~\(S_O\).

\section{METHOD}

In this section, we present our approach to LLM-guided contact point selection, ConPoSe, along with all other steps required for the object pushing task, see Fig.~\ref{fig:Overview}.

\subsection{Object Path Planning}
\label{sec:obj_path_planning}

At the start of the execution, we compute the object path~\(S_O\) using the A* pathfinding algorithm.
The path is optimized on a costmap that balances path length with clearance from obstacles, since the object may deviate from the planned trajectory due to unmodeled object dynamics.
In addition to the untraversable regions around obstacles, we introduce a smooth cost gradient that extends up to \(1.5r_{\text{safety}}\) and cost decreases with increasing distance to obstacles.

To determine linear line segments of consistent direction along the path~\(S_O\), we apply the Ramer-Douglas-Peucker algorithm. 
The resulting simplified path~\(S_O^\text{simplified} = \{\mathbf{x}_{\text{WP}}^1, \dots, \mathbf{x}_{\text{WP}}^{K}\}\) consists of \(K\)~key waypoints (WP). 
We determine the current target pushing direction~\(\varphi^k(t)\) using these key waypoints, which is computed as
\begin{equation}
    \label{eq:target_dir}
    \varphi^k(t) = \arctantwo(y^k_{\text{WP}} - y_O(t), x^k_{\text{WP}} - x_O(t))
\end{equation}
with the next key waypoint toward the goal~\(\mathbf{x}^k_{\text{WP}} = (x_O^k, y_O^k)\). 
The target pushing direction guides the pushing configuration selection step (see Sec.~\ref{sec:push_config_selection}).  

\subsection{Contact Point Generation}
\label{sec:cp_generation}

Since we do not hand-pick robot-object contact points, the set of possible pushing configurations is, in principle, infinite. 
This makes the selection of suitable pushing configurations challenging, as computation becomes intractable. 
Inspired by Tang\etal~\cite{tang2024collaborativeplanarpushingpolytopic}, we address this challenge by automatically generating a finite set of \(M\)~candidate contact points in the object frame at the beginning of each experiment. 
To this end, we discretize the object boundary~\(\delta \mathcal{B}\) into segments of a minimum width of~\(w_{segment} \ge w_{min}\). 
We adjust~\(w_{segment}\) for each edge to achieve equidistant spacing of contact points along the respective edge.  
We introduce additional safety margins near concave corners to guarantee that robots can be placed reliably at every contact point.
For circular objects, whose contour is approximated by many short edges, adjacent edges are aggregated until the target segment width~\(w_{segment}\) is reached. 
The center of each segment is then designated as a candidate contact point~\(c_m\).
For every contact point, we also compute the pushing direction~\(\theta^m_\text{CP}\), defined as the inward-pointing boundary normal.  
The resulting finite set of \(M\)~candidate contact points~\(C = \{c_1,\dots,c_{M}\}\) serves as the pool from which contact points are chosen during pushing configuration selection.

\subsection{ConPoSe: Pushing Configuration Selection}
\label{sec:push_config_selection}

For the selection of contact points, and thus a pushing configuration, we combine the common-sense reasoning capabilities of LLMs with the strengths of local search. 
In our method ConPoSe, we query an LLM~\(\mathcal{L}\) to select an initial candidate pushing configuration~\(\mathcal{P}^z\) as the starting point for local search, see Alg.~\ref{alg:local_search}.
Our zero-shot text prompt~\(l^z_{\text{LLM}}\) contains three components: (i)~a concise task description, (ii)~the current target pushing direction~\(\varphi^k(t)\) (Eq.~\ref{eq:target_dir}) toward the next key waypoint~\(k\) (see Sec.~\ref{sec:obj_path_planning}), and (iii)~the current positions~\(\mathbf{x}_{\text{CP}}^m(t)\) and pushing directions~\(\theta_{\text{CP}}^m(t)\) (in radians) of all contact points transformed into the world frame. 
The LLM is instructed to return a list of distinct contact points, so that the object is pushed as closely as possible along the target direction while minimizing object rotation. 
The complete prompt is available online. 
To ensure consistency, we request structured outputs from the LLM, formulated as 
\begin{equation}
\label{eq:llm_query}
\mathcal{S}^z, \mathcal{P}^z = \mathcal{L}(l^z_{\text{LLM}}),
\end{equation}
where \(\mathcal{S}^z\) is a string with the LLM's reasoning and \(\mathcal{P}^z\) is an array of integers specifying the selected contact point indices.

The candidate pushing configuration is evaluated against two feasibility conditions. 
First, we check for directional alignment.
The resulting pushing direction must lie within an angular tolerance~\(\epsilon\) of the current target pushing direction~\(\varphi^k(t)\) (Eq.~\ref{eq:target_dir}). 
We calculate~\(\epsilon\) as 
\begin{equation}
    \epsilon(t) = \arctantwo(d_{\text{repl}}, d^k_{\text{WP}}(t)),
    \label{equ:epsilon}
\end{equation}
where \(d_{\text{repl}}\) is the maximum allowed path deviation before re-selection is triggered (see Sec.~\ref{sec:online_adaptation}) and \(d^k_{\text{WP}}\) is the Euclidean distance to the next key waypoint. 
In this way, we ensure that the configuration can, in principle, push the object along the entire path segment defined by key waypoint~\(k\). 
Second, we check for sufficient pushing force. 
We assume that the pushing force of at least~\(\frac{N}{2}\)~robots is required to push the object. 
If these conditions are met, the LLM-generated pushing configuration is directly used. 

Otherwise, local search explores nearby alternatives. 
We construct a neighborhood~\(\mathcal{N}\) by randomly selecting a contact point from the current pushing configuration and replacing it with every other feasible contact point from the set of candidate contact points~\(C\). 
Each neighbor is then evaluated whether its resulting pushing direction is closer to the target direction~\(\varphi^k(t)\) than the current best solution and whether it meets the minimum force constraint.
Among all neighbors, we retain the best configuration found so far. 
Search terminates once a configuration within the tolerance~\(\epsilon\) is found or after a maximum of \(I_{\text{max}}\) iterations.
Consequently, the algorithms terminates latest after evaluating~\(I_\text{max} * (M - N) + 1\) pushing configurations.

\begin{algorithm}[t]
\caption{ConPoSe}\label{alg:local_search}
\begin{algorithmic}[1]
    \Require Set of contact points~\(C\), contact point torques~\(\mathbf{\tau}_{m}\), target pushing direction~\(\mathbf{F}^z = (\cos(\varphi^k(t)), \sin(\varphi^k(t)))\),  max. search iterations $I_{\text{max}}$
    \Ensure Pushing configuration~\(\mathcal{P}^z\)
    \State $\mathcal{P}^z = \mathcal{L}(l^z_\text{LLM})$      \Comment{LLM-based initialization}
    \State \rule{0pt}{1.2em}$\mathbf{F}_\text{init} = \sum_{n=0}^{N-1}{ (\cos(\hat{\theta}_\text{init}^n), \sin(\hat{\theta}_\text{init}^n))} |\mathbf{F}_\text{Bot}|$ \\ \Comment{pushing direction}
    \State $\Delta \varphi_\text{best} = \arccos(\mathbf{F}^z \cdot  \mathbf{F}_\text{init})$  \Comment{vector similarity}
    \State $\mathbf{\tau}_\text{best} = \sum_{n=0}^{N-1}{\mathbf{\tau}^{n}_\text{CP}}$ \Comment{torque}
    \If{$\Delta \varphi_\text{best} < \epsilon(t)$ \textbf{and} $|\mathbf{\mathbf{F}_\text{init}}|  > \frac{N}{2} |\mathbf{F}_\text{Bot}|$}
        \State \Return $\mathcal{P}^z$
    \EndIf
    \For{$i=0$ \textbf{to} $I_{\text{max}}$} \Comment{local search}
        \State $\mathcal{N} = \textsc{ReplaceOne}(\mathcal{P}^z, C)$ \Comment{neighborhood}
        \For{$\mathcal{P}_{\text{cand}}\in\mathcal{N}$}
            \State $\mathbf{F}_\text{cand} = \sum_{n=0}^{N-1}{ (\cos(\hat{\theta}_\text{cand}^n), \sin(\hat{\theta}_\text{cand}^n))|\mathbf{F}_\text{Bot}}|$
            \State $\Delta \varphi_\text{cand} = \arccos(\mathbf{F}^z \cdot  \mathbf{F}_\text{cand})$
            \State $\mathbf{\tau}_\text{cand} = \sum_{n=0}^{N-1}{\mathbf{\tau}_{n}}$
            \If{$\Delta \varphi_\text{cand} < \epsilon(t)$ \textbf{and} $|\mathbf{\mathbf{F}_\text{cand}}|  > \frac{N}{2} |\mathbf{F}_\text{Bot}|$}
                \State \Return $\mathcal{P}_{\text{cand}}$
            \EndIf
            \If{$(\Delta \varphi_\text{cand}> \Delta\varphi_{best}$ \textbf{or}\par 
                \hskip\algorithmicindent$\Delta \varphi_\text{cand}=\Delta\varphi_{best}$ \textbf{and} $\tau < \tau_\text{best})$ \textbf{and}\par 
                \hskip\algorithmicindent$|\mathbf{\mathbf{F}_\text{cand}}|  > \frac{N}{2} |\mathbf{F}_\text{Bot}|$}
                \State $\mathcal{P}^z = \mathcal{P}_\text{cand}$
                \State $\Delta \varphi_\text{best} = \Delta \varphi_\text{cand}$
                \State $\mathbf{\tau}_\text{best} = \mathbf{\tau}_\text{cand}$
            \EndIf
        \EndFor
    \EndFor
    \State \Return \(\mathcal{P}^z\) 
\end{algorithmic}
\end{algorithm}

\subsection{Online Adaptation}
\label{sec:online_adaptation}

Since we do not model the physics of the object-robot-environment interaction, our approach does not explicitly account for friction and uncertainties such as slipping. 
Consequently, the actual pushing direction may deviate from the direction determined in the pushing configuration selection step. 
To ensure robot progress toward the goal despite the possible deviations, we define three conditions under which the pushing configuration is re-selected: (i)~the next key waypoint~\(k\) was reached, (ii)~the object has deviated from the planned object path~\(S_O\) by more than a threshold distance~\(d_\text{repl}\), (iii)~the object has not made measurable progress towards the goal in~\(t_\text{repl}\).

\subsection{Contact Point Switching}
\label{sec:cp_switching}

We adopt the robot-contact point mapping approach from Tang\etal~\cite{tang2024collaborativeplanarpushingpolytopic}. 
The robot positions~\(\mathbf{x}_\text{Bot}\) and the contact points of the selected pushing configuration~\(\mathcal{P}^z\) are projected onto a circle~$\mathcal{C}$ centered at the object, preserving the relative ordering of the positions.
Next, we select an angle $\alpha^*\in[0,\pi]$, which splits the circle $\mathcal{C}$ into two semicircles $\mathcal{C^+}$ and $\mathcal{C^-}$, such that the number of contact points and robots is the same on each semicircle. 
Robots and contact points are numbered clockwise from~\(\alpha^*\) and each robot is then assigned to the contact point with the same index. 

Before switching contact points, the robots realign themselves with their current contact point.
Afterward, a local path to the newly assigned contact point~\(\hat{c}^n\) is planned for each robot using the A* algorithm. 
The contact point matching algorithm ensures, under free movement around the object, that no inter-robot collisions occur and individual paths are bounded by at most half of the object's perimeter. 
Based on this property, we allow robots to switch contact points concurrently when all planned paths are shorter than half of the object perimeter and maximum one robot is located close to a concave corner. 
Otherwise, robots perform the switch consecutively to avoid collisions. 
We use an adaptive pure pursuit algorithm for path tracking~\cite{coulter1992implementation}. 

\subsection{Pushing}
\label{sec:pushing}

We use P-controllers to generate the control inputs for each robot during pushing.
Since maintaining a stable robot-object contact point with differential drive robots is difficult, we prioritize maintaining contact with the object boundary over precisely preserving the assigned contact point position and pushing direction.  
The linear velocity of each robot~\(n\) is calculated by 
\begin{equation}
    v^n(t) = k_v(\min(\lVert \delta B - \mathbf{x}^n(t))\rVert_2), \, k_v > 0.
\end{equation}
The angular velocity is calculated by
\begin{multline}
    \omega^n(t) = k_{\omega_1} (\hat{\theta}_\text{CP}^n(t) - \theta^n_\text{Bot}(t)) + k_{\omega_2}(\beta_\text{CP}(t) - \theta^n_\text{Bot}(t)), \\ 
k_{\omega_1}, k_{\omega_2} > 0, 
\end{multline}
where \(\beta_\text{CP}(t) = \arctantwo(\hat{y}^n_\text{CP}(t) - y^n(t), \hat{x}_{CP}^n(t) - x^n(t))\) is the heading toward the assigned contact point~\(\hat{c}_n\).
To prevent robots from pushing the object in unintended directions while following curved trajectories, we let robots alternate quickly between turning and linear motion steps.

\section{EXPERIMENTS}

\begin{figure}
     \centering
     \includegraphics[width=0.49\linewidth]{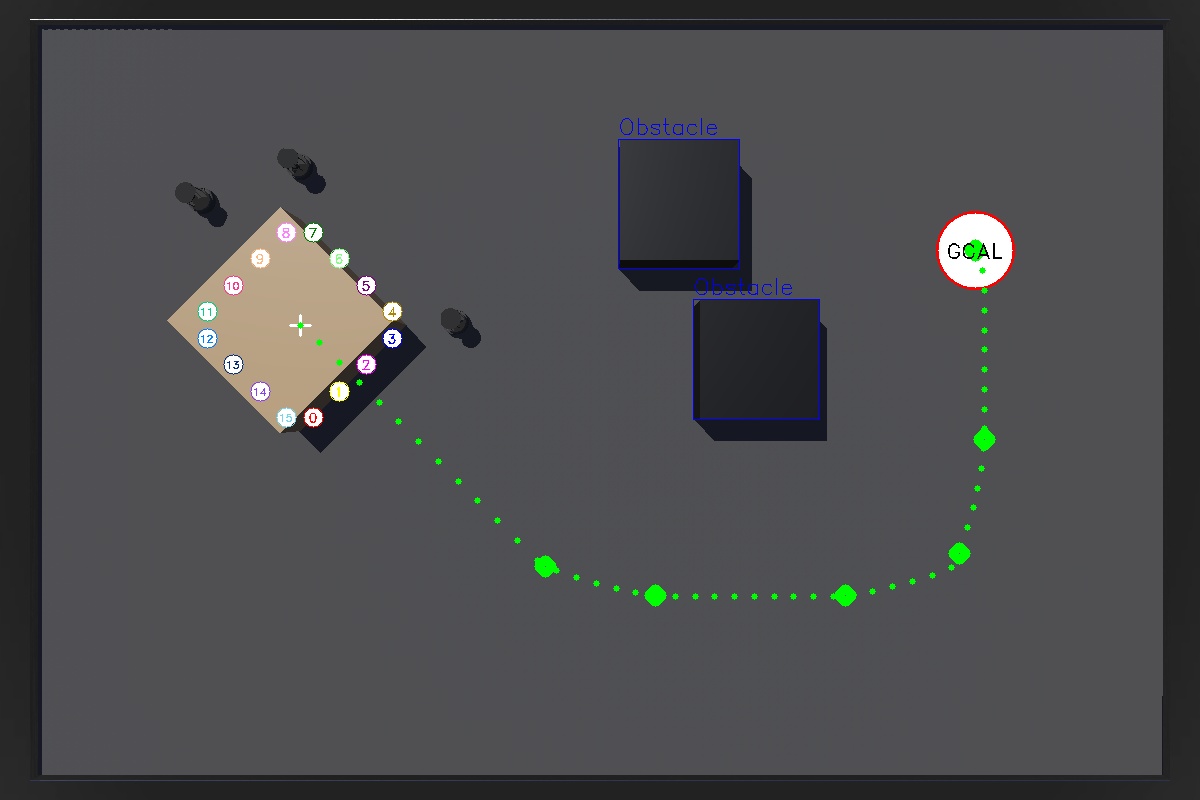}
     \includegraphics[width=0.49\linewidth]{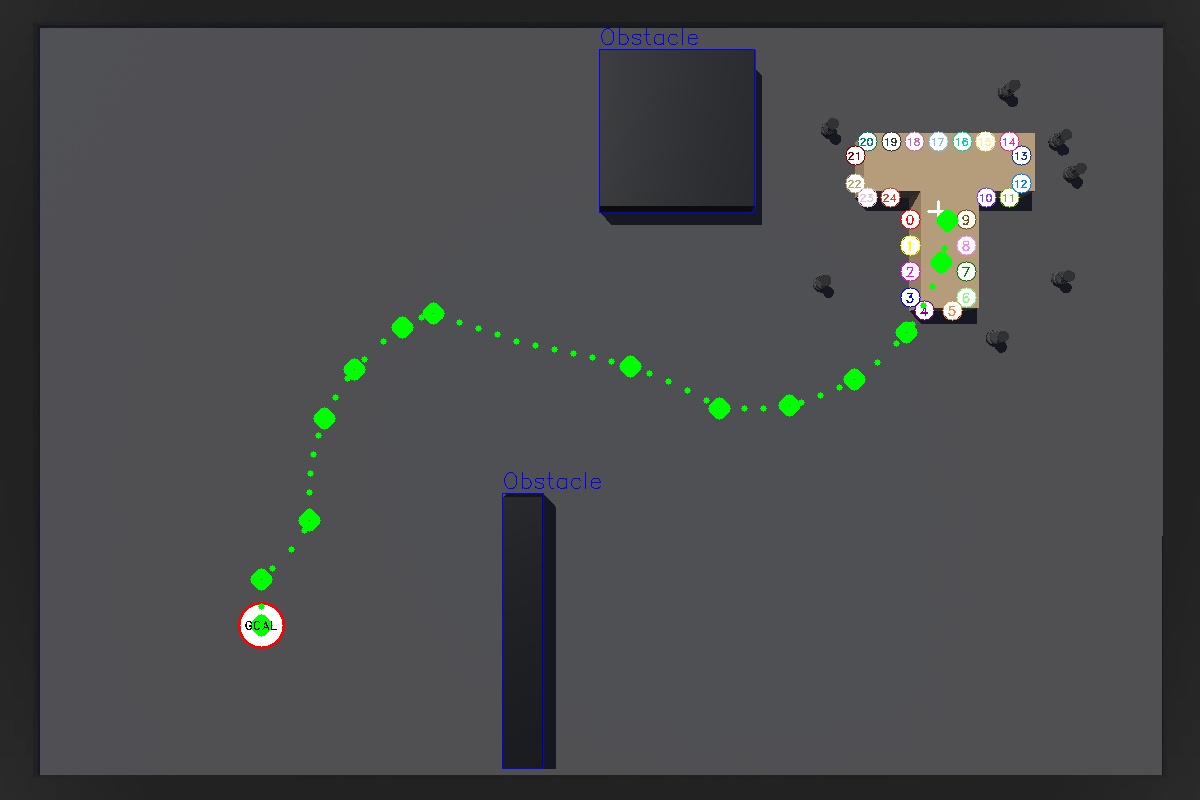} \\[0.015\linewidth]
     \includegraphics[width=0.49\linewidth]{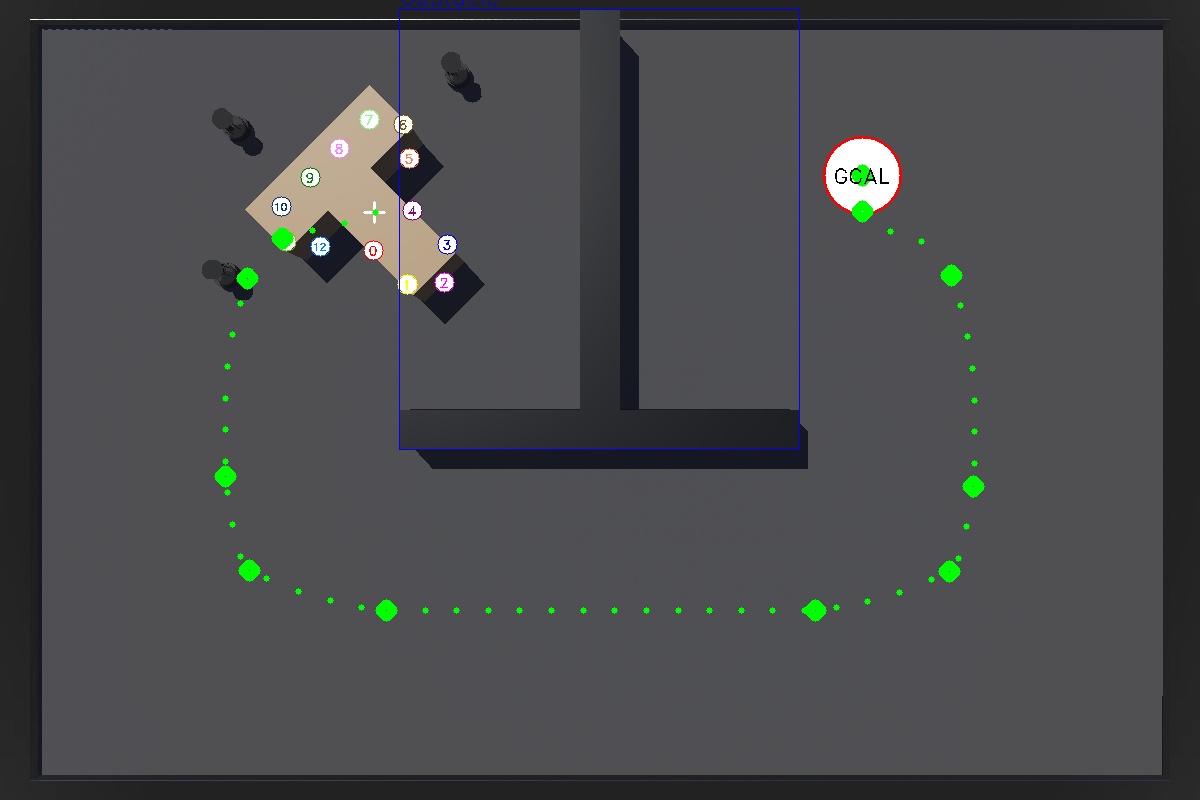}
     \includegraphics[width=0.49\linewidth]{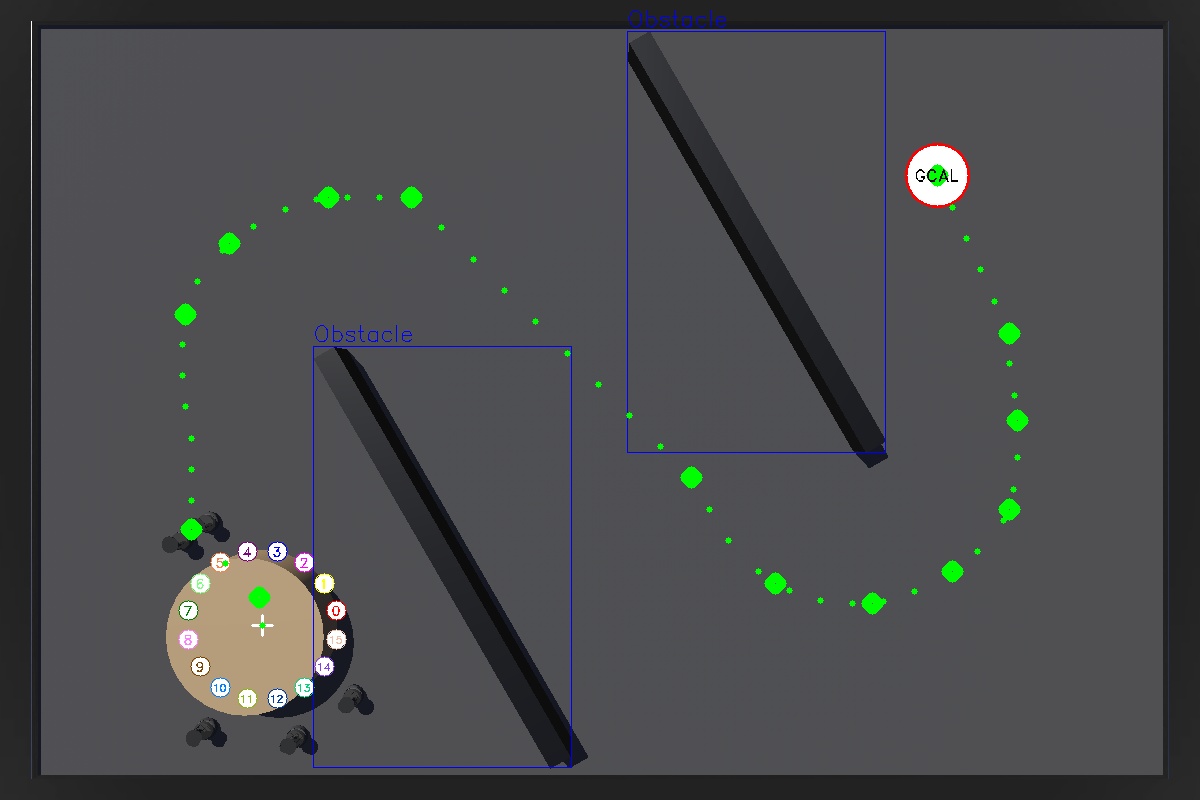}
     \caption{Examples of four of our five scenes (the remaining scene is shown in Fig.~\ref{fig:Overview}). For illustration purposes, each image is annotated with the planned object path~\(S_O\) (green dots), its simplified representation~\(S_O^\text{simplified}\) (green diamonds), the goal position, the object center (white plus sign), and the \(M\)~candidate contact points (numbered circles).}
     \label{fig:annotated_image}
 \end{figure}

\subsection{Setup}

We conduct experiments in the Webots simulator~\cite{Webots04} with homogeneous MRSs of \(N \in \{ 3, 5, 7  , 10, 12, 15\}\) TurtleBot~4 robots. 
We set a maximum linear velocity of~\(0.2\,\frac{\text{m}}{\text{s}}\) for pushing and of~\(0.1\,\frac{\text{m}}{s}\) for contact point switching, and a maximum angular velocity of~\(1.0\,\frac{\text{rad}}{\text{s}}\). 
Experiments are performed in a \(10\,\text{m} \times 15\,\text{m}\)~arena across five scenes with varying obstacle constellations and object start and goal positions, see Fig.~\ref{fig:annotated_image}.
We test three object shapes: (i)~a cuboid of \(2.0\,\text{m} \times 2.0\,\text{m}\), (ii)~a cylinder with a radius of \(1.0\,\text{m}\), and (iii)~a T-Shape with a horizontal bar of~\(2.2\,\text{m}\times 0.75\,\text{m}\) and a vertical bar of~\(0.75\, \text{m} \times 1.5\, \text{m}\). 
All objects have a height of~\(0.5\,\text{m}\) and a density of~\(1.0~\frac{\text{kg}}{\text{m}^3} \).
For \(N > 4\), arena, obstacle, and object dimensions are scaled by a factor of~\(\frac{N}{4}\) to ensure feasible pushing configurations.
Robot–object and ground friction coefficients are set to 0.5 and 1.0, respectively. 
We consider the goal reached if the object's center is within \(0.5\,\text{m}\) of the goal position.
For ConPoSe~(Alg.~\ref{alg:local_search}), we set the number of maximum search iterations~\(I_\text{max}\) to~\(5\). 
All simulations are conducted on an AMD Ryzen 9 7900 CPU with an NVIDIA GeForce RTX 4080 GPU.

\subsection{Baselines}

We compare ConPoSe against a naive LLM-based and an analytical baseline in our experiments. 

\paragraph{Naive LLM-based Baseline (N-LLM)}
\label{sec:llm-selection}

To demonstrate how local search refines the initial candidate generated by the LLM, we compare ConPoSe with a purely LLM-based contact point selection method. 
In this baseline, the raw LLM output (see Eq.~\ref{eq:llm_query}) is used directly as the pushing configuration, without applying any subsequent optimization or refinement.  

\begin{figure*}[t]
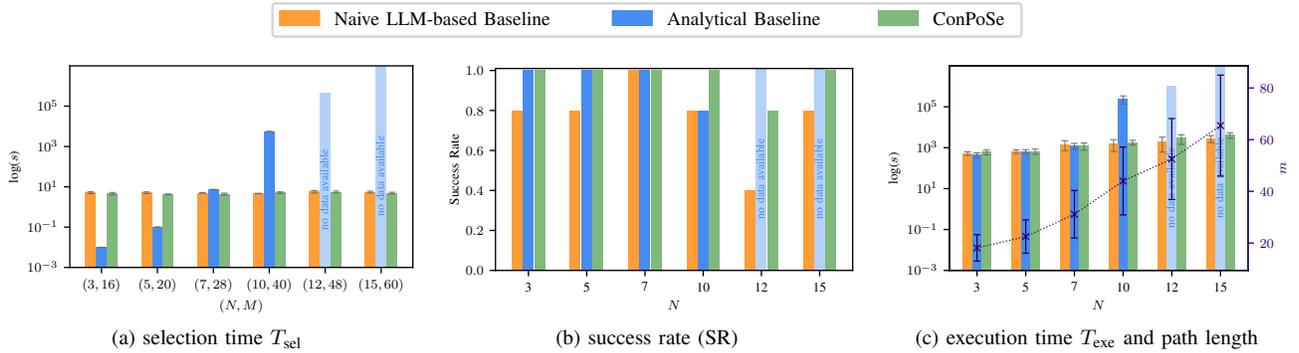

    \centering 
    \begin{adjustbox}{clip,trim=0cm 5.15cm 0cm 5.15cm,max width=0.9\linewidth}
    %% Creator: Matplotlib, PGF backend
%%
%% To include the figure in your LaTeX document, write
%%   \input{<filename>.pgf}
%%
%% Make sure the required packages are loaded in your preamble
%%   \usepackage{pgf}
%%
%% Also ensure that all the required font packages are loaded; for instance,
%% the lmodern package is sometimes necessary when using math font.
%%   \usepackage{lmodern}
%%
%% Figures using additional raster images can only be included by \input if
%% they are in the same directory as the main LaTeX file. For loading figures
%% from other directories you can use the `import` package
%% and then include the figures with
%%   \import{<path to file>}{<filename>.pgf}
%%
%% Matplotlib used the following preamble
%%   \def\mathdefault#1{#1}
%%   \everymath=\expandafter{\the\everymath\displaystyle}
%%   \IfFileExists{scrextend.sty}{
%%     \usepackage[fontsize=8.000000pt]{scrextend}
%%   }{
%%     \renewcommand{\normalsize}{\fontsize{8.000000}{9.600000}\selectfont}
%%     \normalsize
%%   }
%%   
%%   \makeatletter\@ifpackageloaded{underscore}{}{\usepackage[strings]{underscore}}\makeatother
%%
\begingroup%
\makeatletter%
\begin{pgfpicture}%
\pgfpathrectangle{\pgfpointorigin}{\pgfqpoint{6.918500in}{4.275868in}}%
\pgfusepath{use as bounding box, clip}%
\begin{pgfscope}%
\pgfsetbuttcap%
\pgfsetmiterjoin%
\definecolor{currentfill}{rgb}{1.000000,1.000000,1.000000}%
\pgfsetfillcolor{currentfill}%
\pgfsetlinewidth{0.000000pt}%
\definecolor{currentstroke}{rgb}{1.000000,1.000000,1.000000}%
\pgfsetstrokecolor{currentstroke}%
\pgfsetdash{}{0pt}%
\pgfpathmoveto{\pgfqpoint{0.000000in}{0.000000in}}%
\pgfpathlineto{\pgfqpoint{6.918500in}{0.000000in}}%
\pgfpathlineto{\pgfqpoint{6.918500in}{4.275868in}}%
\pgfpathlineto{\pgfqpoint{0.000000in}{4.275868in}}%
\pgfpathlineto{\pgfqpoint{0.000000in}{0.000000in}}%
\pgfpathclose%
\pgfusepath{fill}%
\end{pgfscope}%
\begin{pgfscope}%
\pgfsetbuttcap%
\pgfsetmiterjoin%
\definecolor{currentfill}{rgb}{1.000000,1.000000,1.000000}%
\pgfsetfillcolor{currentfill}%
\pgfsetfillopacity{0.800000}%
\pgfsetlinewidth{1.003750pt}%
\definecolor{currentstroke}{rgb}{0.800000,0.800000,0.800000}%
\pgfsetstrokecolor{currentstroke}%
\pgfsetstrokeopacity{0.800000}%
\pgfsetdash{}{0pt}%
\pgfpathmoveto{\pgfqpoint{1.328750in}{2.042490in}}%
\pgfpathlineto{\pgfqpoint{5.589750in}{2.042490in}}%
\pgfpathquadraticcurveto{\pgfqpoint{5.611972in}{2.042490in}}{\pgfqpoint{5.611972in}{2.064712in}}%
\pgfpathlineto{\pgfqpoint{5.611972in}{2.211156in}}%
\pgfpathquadraticcurveto{\pgfqpoint{5.611972in}{2.233378in}}{\pgfqpoint{5.589750in}{2.233378in}}%
\pgfpathlineto{\pgfqpoint{1.328750in}{2.233378in}}%
\pgfpathquadraticcurveto{\pgfqpoint{1.306528in}{2.233378in}}{\pgfqpoint{1.306528in}{2.211156in}}%
\pgfpathlineto{\pgfqpoint{1.306528in}{2.064712in}}%
\pgfpathquadraticcurveto{\pgfqpoint{1.306528in}{2.042490in}}{\pgfqpoint{1.328750in}{2.042490in}}%
\pgfpathlineto{\pgfqpoint{1.328750in}{2.042490in}}%
\pgfpathclose%
\pgfusepath{stroke,fill}%
\end{pgfscope}%
\begin{pgfscope}%
\pgfsetbuttcap%
\pgfsetmiterjoin%
\definecolor{currentfill}{rgb}{1.000000,0.615686,0.215686}%
\pgfsetfillcolor{currentfill}%
\pgfsetlinewidth{1.003750pt}%
\definecolor{currentstroke}{rgb}{1.000000,0.615686,0.215686}%
\pgfsetstrokecolor{currentstroke}%
\pgfsetdash{}{0pt}%
\pgfpathmoveto{\pgfqpoint{1.350972in}{2.109601in}}%
\pgfpathlineto{\pgfqpoint{1.573194in}{2.109601in}}%
\pgfpathlineto{\pgfqpoint{1.573194in}{2.187378in}}%
\pgfpathlineto{\pgfqpoint{1.350972in}{2.187378in}}%
\pgfpathlineto{\pgfqpoint{1.350972in}{2.109601in}}%
\pgfpathclose%
\pgfusepath{stroke,fill}%
\end{pgfscope}%
\begin{pgfscope}%
\definecolor{textcolor}{rgb}{0.000000,0.000000,0.000000}%
\pgfsetstrokecolor{textcolor}%
\pgfsetfillcolor{textcolor}%
\pgftext[x=1.662083in,y=2.109601in,left,base]{\color{textcolor}{\rmfamily\fontsize{8.000000}{9.600000}\selectfont\catcode`\^=\active\def^{\ifmmode\sp\else\^{}\fi}\catcode`\%=\active\def%{\%}Naive LLM-based Baseline}}%
\end{pgfscope}%
\begin{pgfscope}%
\pgfsetbuttcap%
\pgfsetmiterjoin%
\definecolor{currentfill}{rgb}{0.274510,0.549020,0.941176}%
\pgfsetfillcolor{currentfill}%
\pgfsetlinewidth{1.003750pt}%
\definecolor{currentstroke}{rgb}{0.274510,0.549020,0.941176}%
\pgfsetstrokecolor{currentstroke}%
\pgfsetdash{}{0pt}%
\pgfpathmoveto{\pgfqpoint{3.263861in}{2.109601in}}%
\pgfpathlineto{\pgfqpoint{3.486083in}{2.109601in}}%
\pgfpathlineto{\pgfqpoint{3.486083in}{2.187378in}}%
\pgfpathlineto{\pgfqpoint{3.263861in}{2.187378in}}%
\pgfpathlineto{\pgfqpoint{3.263861in}{2.109601in}}%
\pgfpathclose%
\pgfusepath{stroke,fill}%
\end{pgfscope}%
\begin{pgfscope}%
\definecolor{textcolor}{rgb}{0.000000,0.000000,0.000000}%
\pgfsetstrokecolor{textcolor}%
\pgfsetfillcolor{textcolor}%
\pgftext[x=3.574972in,y=2.109601in,left,base]{\color{textcolor}{\rmfamily\fontsize{8.000000}{9.600000}\selectfont\catcode`\^=\active\def^{\ifmmode\sp\else\^{}\fi}\catcode`\%=\active\def%{\%}Analytical Baseline}}%
\end{pgfscope}%
\begin{pgfscope}%
\pgfsetbuttcap%
\pgfsetmiterjoin%
\definecolor{currentfill}{rgb}{0.498039,0.725490,0.486275}%
\pgfsetfillcolor{currentfill}%
\pgfsetlinewidth{1.003750pt}%
\definecolor{currentstroke}{rgb}{0.498039,0.725490,0.486275}%
\pgfsetstrokecolor{currentstroke}%
\pgfsetdash{}{0pt}%
\pgfpathmoveto{\pgfqpoint{4.792639in}{2.109601in}}%
\pgfpathlineto{\pgfqpoint{5.014861in}{2.109601in}}%
\pgfpathlineto{\pgfqpoint{5.014861in}{2.187378in}}%
\pgfpathlineto{\pgfqpoint{4.792639in}{2.187378in}}%
\pgfpathlineto{\pgfqpoint{4.792639in}{2.109601in}}%
\pgfpathclose%
\pgfusepath{stroke,fill}%
\end{pgfscope}%
\begin{pgfscope}%
\definecolor{textcolor}{rgb}{0.000000,0.000000,0.000000}%
\pgfsetstrokecolor{textcolor}%
\pgfsetfillcolor{textcolor}%
\pgftext[x=5.103750in,y=2.109601in,left,base]{\color{textcolor}{\rmfamily\fontsize{8.000000}{9.600000}\selectfont\catcode`\^=\active\def^{\ifmmode\sp\else\^{}\fi}\catcode`\%=\active\def%{\%}ConPoSe}}%
\end{pgfscope}%
\end{pgfpicture}%
\makeatother%
\endgroup%
    \end{adjustbox}\\
    \subfloat[selection time~\(T_\text{sel}\)]{
        \resizebox{.315\linewidth}{!}{
        \begin{adjustbox}{clip,trim=0cm 0cm 0cm 0.1cm,max width=\textwidth}
        \input{img/scalability.pgf}
        \end{adjustbox}}
         \label{fig:st_scaling}
         } 
    \subfloat[success rate (SR)]{
        \resizebox{.315\linewidth}{!}{ \begin{adjustbox}{clip,trim=0cm 0cm 0cm 0.1cm,max width=\textwidth}
         \input{img/success_rate.pgf}
        \end{adjustbox}}
         \label{fig:sr_scaling}
         }
    \subfloat[execution time~\(T_\text{exe}\) and path length]{
       \resizebox{.315\linewidth}{!}{ \begin{adjustbox}{clip,trim=0cm 0cm 0cm 0.1cm,max width=\textwidth}
         \input{img/time.pgf}
        \end{adjustbox}}
         \label{fig:time_scaling}
         }
  \caption{Scalability analysis: comparison of selection time~\(T_{sel}\), success rate (SR), and execution time (\(T_{exe}\)) for \(N \in \{3, 5, 7, 10, 12, 15\}\). Data for the baseline is limited to \(N < 12\) due to excessively long execution times.}
  \label{fig:query-time-bar}
\end{figure*}

\paragraph{Analytical Baseline (A-BL)}

As a second baseline, we analytically determine the best pushing configuration. 
Given the current target pushing direction~\(\varphi^k(t)\) (Eq.~\ref{eq:target_dir}), along with the candidate contact point positions~\(\mathbf{x}^m_{\text{CP}}(t)\) and their associated pushing directions~\(\theta^m_{\text{CP}}\), we evaluate all possible configurations. 
Specifically, we consider all \(G=\binom{M}{N}\)~combinations of \(N\)~distinct contact points from the \(M\)~available candidates. 
For each pushing configuration, we calculate the resulting pushing direction by summing the individual robot pushing vectors. 
From these, we select the configuration that (i)~produces a direction closest to the target~\(\varphi^k(t)\), (ii)~achieves a minimum force magnitude equal to~\(\frac{N}{2}\)~robots pushing in the target direction, and (iii)~minimum torque~\(\tau\), thereby reducing undesired object rotation. 
The complete algorithm is outlined in Alg.~\ref{alg:baseline}. 
The primary limitation of this baseline is its exponential time complexity~\(\mathcal{O}(2^M)\), which renders it computationally impractical for large numbers of candidate contact points and robots. 

\begin{algorithm}[t]
\caption{Analytical Baseline}\label{alg:baseline}
\begin{algorithmic}[1]
    \Require Set of contact points~\(C\), contact point torques~\(\mathbf{\tau}_{m}\), target pushing direction~\(\mathbf{F}^z = (\cos(\varphi^k(t)), \sin(\varphi^k(t))\) 
    \Ensure Pushing configuration~\(\mathcal{P}^k_z\)
    \State $\mathcal{P}^z = \text{None}$     \Comment{initialization}
    \For{$\mathcal{P}_\text{cand} \in \binom{C}{N}$} \Comment{every combination}
    \State $\mathbf{F}_\text{cand} = \sum_{n=0}^{N-1}{ (\cos(\hat{\theta}_\text{cand}^n), \sin(\hat{\theta}_\text{cand}^n)) |\mathbf{F}_\text{Bot}|}$ \\ \Comment{pushing direction}
    \State $\Delta \varphi_\text{cand} = \text{arccos}(\mathbf{F}^z \cdot  \mathbf{F}_\text{cand})$  \Comment{vector similarity}
    \State{$\mathbf{\tau}_\text{cand} = \sum_{n=0}^{N-1}{\mathbf{\tau}^{n}_\text{cand}}$} \Comment{torque}
    \If{$\Delta \varphi_\text{cand} < \Delta\varphi_\text{best} $ \textbf{and} $|\mathbf{\mathbf{F}_\text{cand}}|  > \frac{N}{2} |\mathbf{F}_\text{Bot}|$ \textbf{and} \par
        \hskip\algorithmicindent  $\tau < \tau_\text{best} $}
     \State $\mathcal{P}^z = \mathcal{P}_\text{cand}$
    \EndIf
    \EndFor   
\State \Return \(\mathcal{P}^z\) 
\end{algorithmic}
\end{algorithm}

\subsection{Evaluation Metrics}

For our evaluations, we report one or more of the following five metrics:
\begin{LaTeXdescription}
    \item[Success Rate (SR):] proportion of trials in which the object reaches the goal.
    \item[Number of pushing configurations~\(Z\):] total number of pushing configurations selected per run.
    \item[Selection time~(\(T_{sel}\)):] mean time required to select one pushing configuration. 
    \item[Execution time~(\(T_{exe}\)):] total experiment duration from start to finish, including all steps shown in Fig.~\ref{fig:Overview}. 
    \item[Switch time~(\(T_{sw}\)):] mean time required to execute contact point switching from one pushing configuration to the next.  
\end{LaTeXdescription}

\subsection{LLM Comparison}

\begin{table}
    \centering
    \caption{Comparison of different LLMs}
    \label{tab:comparison_VLM}
     \begin{tabular}{lccc}
      \toprule
      Model & SR (\(\uparrow\)) & \(T_{sel}\) [\(\text{s}\)] (\(\downarrow\)) & \(Z\) [\#] (\(\downarrow\))  \\
      \midrule
      GPT-4.1 & \(0.8\) & \(\mathbf{(5.12\pm0.53)}\) & \(19.60 \pm 3.14\)  \\ 
     \rowcolor{SeaGreen3!10!} GPT-5 & \(\mathbf{1.0}\) & \((127.87\pm18.86)\) & \(\mathbf{17.60 \pm 5.95}\)  \\ 
     o4-mini & \(0.8\) & \((33.85\pm1.76)\) & \(18.40  \pm 5.85 \)  \\ 
      \rowcolor{SeaGreen3!10!} Gemma~3 & \(0.0\) &\((18.40\pm9.42)
      \)& \(8.60 \pm 4.22\) \\ 
       LLaVa & \(0.0\) & \((3.56\pm2.37)\) & \(3.40 \pm 2.15 \) \\  
      \midrule
       \rowcolor{SeaGreen3!10!} A-BL & \(1.0\) & \(0.01\) & \(19.00 \pm 9.08\) \\
      \bottomrule
     \end{tabular}
\end{table}

As a first step, we evaluate different LLMs using our naive LLM-based baseline to select the model for the subsequent experiments. 
We compare GPT-4.1, GPT-5, o4-mini, Gemma~3 (4.3B), and LLaVa (7B). 
Each model is tested across the five scenes with \(N=3\)~robots and the cuboid object. 
For reference, we also report the results of the analytical baseline. 

As shown in Tab.~\ref{tab:comparison_VLM}, Gemma~3 and LLaVa fail to complete the task.  
Among the OpenAI models, GPT-5 achieves the highest success rate, though it comes at the cost of significantly longer selection times. 
GPT-4.1 offers the best trade-off: it combines fast selection with high success rates, requiring only slightly more pushing configurations than GPT-5 and o4-mini, while remaining close to the analytical baseline.
Based on these results, we select GPT-4.1 for both ConPoSe and the naive LLM-based baseline. 
We expect the local search component of ConPoSe to improve the success rate while maintaining short selection times.

\subsection{Pushing Configuration Selection Evaluation}

Next, we evaluate ConPoSe (see Sec.~\ref{sec:push_config_selection}) and the two baselines under the assumption that robots are automatically repositioned at their newly assigned contact points. 
This setup ensures a fair comparison of the contact point selection methods by preventing failures caused by our contact point switching method. 
A comprehensive evaluation of the full approach is presented in Sec.~\ref{sec:full_system}.

\subsubsection{Scalability}
\label{sec:scalability}

We evaluate ConPoSe and our two baselines using different MRS sizes to assess scalability. 
Each method is tested once across the five scaled scenes with \(N= \{3, 5, 7, 10, 12, 15\}\)~robots using cuboid objects.

\paragraph*{Selection Time} 
Selection time does not increase with the number of robots and contact points for our method, ConPoSe, and the naive LLM-based baseline, see Fig.~\ref{fig:st_scaling}. 
Despite its additional computations, ConPoSe achieves selection times close to the naive LLM-based baseline.
In some cases, the selection time of ConPoSe is even lower than that of the naive LLM-based baseline, which we attribute to varying server response times. 
As expected, the analytical baseline shows exponentially growing selection times, with computations for a single pushing configuration taking more than an hour for \(N=10\)~robots and several days for \(N=12\)~robots.

\paragraph*{Success Rate}

As shown in Fig.~\ref{fig:sr_scaling}, ConPoSe consistently achieves high success rates across all MRS sizes. 
For \(N=12\)~robots, a single run failed, caused by the LLM outputting an incorrect number of contact points. 
Similarly, the analytical baseline reaches high success rates for all~\(N\). 
We observed only one failure at \(N=10\), where the object did not make significant progress toward the goal position for ten consecutive pushing configurations. 
However, we cannot report success rates for~\(N>10\), as the experiments could not be completed within a reasonable time due to the long selection times. 
In contrast, the naive LLM-based baseline exhibits fluctuating success rates, though it remains capable of completing tasks up to~\(N=15\)~robots. 
Here, the primary failure reason is again the LLM generating an incorrect number of contact points.

\paragraph*{Execution Time}
Since the scene size scales with the number of robots, the object path length also increases with MRS size, see Fig.~\ref{fig:time_scaling}. 
Object path lengths are identical across all methods, so we expect execution time to grow with the number of robots for ConPoSe and the two baselines. 
For the analytical baseline,  execution time increases rapidly, driven mainly by the increasing selection time. 
In contrast, ConPoSe and the naive LLM-based baseline show only a moderate increase, which is consistent with expectations: as path length grows, the chance of re-selection due to path deviations grows. 
In addition, the number of key waypoints increases with path lengths, and thus number of robots, from~\(9.80 \pm 2.93\) for \(N=3\) to~\(19.00 \pm 5.02\) for \(N=15\). 

Overall, ConPoSe proves most scalable, as it maintains high success rates and short selection times across all~\(N\).
From \(N=7\) onward, its selection time is equal or shorter than the analytical baseline while keeping high success rates of \(80\,\%\) to \(100\,\%\).

\subsubsection{Different Object Shapes}

Next, we evaluate ConPoSe and the two baselines on three different object shapes, with each method tested across the five scenes and \(N= \{3, 5, 7\}\)~robots. 

The analytical baseline achieves the highest success rate across all three objects, see Tab.~\ref{tab:shapes}. 
ConPoSe performs similarly well for the cylinder and the cuboid, though slightly worse for the T-shape. 
The lower success rates of the naive LLM-based baseline mainly stem from collisions or invalid contact points.

Again, the analytical baseline reaches the shortest execution time, followed by ConPoSe and the naive LLM-based baseline. 
As seen in Sec.~\ref{sec:scalability}, ConPoSe becomes competitive for \(N \ge 7\) since the selection time of the baseline increases exponentially. 
Due to the long execution times of the analytical baseline for higher \(N\), we restricted our comparison here to \(N\le7\) and thus gave the analytical baseline an advantage. 
Based on the previous results, we again expect ConPoSe to outperform the analytical baseline in execution time for higher~\(N\). 

For all objects and methods, the number of pushing configurations exceeds the mean number of key waypoints (\(K = 11.33 \pm 3.24\) for T-shape, \(K = 8.80 \pm 3.37\) for cylinder, \(K = 11.13 \pm 2.99\) for cuboid), which is likely due to the unmodeled slippage and friction. 
The analytical baseline requires the fewest configurations, again followed closely by ConPoSe.

The cylinder emerges as the easiest object to transport, requiring fewer pushing configurations~\(Z\) and less total execution time than the other two objects.
Remarkably, it achieves a \(100\,\%\)~success rate in all three evaluated methods, which is unmatched by the other shapes. 
We attribute this to two reasons: 
First, each contact point corresponds to a unique pushing direction for cylindrical objects, allowing a more fine-grained pushing configuration selection. 
In contrast, angular objects are restricted to a limited number of directions normal to their boundary.  
Second, cylinders are inherently rotation invariant, and any applied pushing force is directed toward the center of mass. 
As a result, the cylinder is less affected by deviations due to unwanted rotations that commonly occur with angular shapes.

\subsection{Full System Evaluation}
\label{sec:full_system}

\begin{table}
    \centering
    \caption{Object shape comparison \label{tab:shapes}}
     \begin{tabular}{clcccccc}
      \toprule
       & Model &  SR (\(\uparrow\)) &  \(T_{exe}\) (\(\downarrow\)) & \(Z\) (\(\downarrow\)) & \(T_{sel}\) (\(\downarrow\)) \\
       & & & [min] & [\#] & [s] \\ 
      \midrule
      & N-LLM & \(0.87\) & \(14.6 \pm 10.0\) & \(27.0 \pm 11.4\) & \(5.1 \pm 0.4\) \\ 
     \rowcolor{SeaGreen3!10!} \cellcolor{white}    & A-BL & \(\mathbf{1.0}\) & \(\mathbf{13.0 \pm 7.1}\) & \(\mathbf{21.1 \pm 8.4}\) & \(\mathbf{2.5 \pm 3.4}\) \\ 
   \multirow{-3}{*}{\rotatebox[origin=c]{90}{Cuboid}}  & \textbf{ConPoSe} & \(\mathbf{1.0}\) & \(13.8 \pm 6.9\) & \(23.9 \pm 7.6\) & \(4.4 \pm 0.5 \) \\ 
      \midrule    
      \rowcolor{SeaGreen3!10!} \cellcolor{white} & N-LLM & \(\mathbf{1.0}\) & \(12.0 \pm 6.3\) & \(23.1 \pm 13.0\) & \(5.0 \pm 0.8\) \\ 
        & A-BL & \(\mathbf{1.0}\) & \(\mathbf{8.7 \pm 3.9}\) & \(\mathbf{15.8 \pm 6.4}\) & \(\mathbf{0.4 \pm 0.5}\)  \\ 
      \rowcolor{SeaGreen3!10!} \cellcolor{white}\multirow{-3}{*}{\rotatebox[origin=c]{90}{Cylinder}} & \textbf{ConPoSe} & \(\mathbf{1.0}\) & \(11.4 \pm 5.5\) &  \(18.0 \pm 7.5\) & \(6.3 \pm 1.4\) \\ 
      \midrule 
       & N-LLM & \(0.67\) & \(\mathbf{10.0 \pm 4.9}\) & \(22.1 \pm 10.2\) & \(4.4 \pm 1.4\) \\ 
        \rowcolor{SeaGreen3!10!} \cellcolor{white}    & A-BL & \(\mathbf{0.93}\) & \(11.7 \pm 5.2\) & \(\mathbf{19.0 \pm 6.6}\) & \(\mathbf{1.0 \pm 1.4}\) \\ 
       \multirow{-3}{*}{\rotatebox[origin=c]{90}{T-Shape}} & \textbf{ConPoSe} & \(\mathbf{0.93}\) & \(13.5 \pm 5.4\) & \(21.7 \pm 8.2\) & \(5.8 \pm 0.8\)  \\ 
      \bottomrule
     \end{tabular}
\end{table}

\begin{table}
    \centering
    \caption{Full system evaluation}
    \label{tab:full_system}
     \begin{tabular}{clcccccc}
      \toprule
       & Model &  SR (\(\uparrow\)) &  \(T_{exe}\)(\(\downarrow\)) & \(Z\) (\(\downarrow\)) & \(T_{sw}\) (\(\downarrow\)) \\
       & & & [min] & [\#] & [min]\\
      \midrule
        & N-LLM & \(\mathbf{0.80}\) & \( 46.2 \pm 25.7 \) & \(28.6 \pm 12.5\) & \(1.5 \pm 0.4\) \\ 
   \rowcolor{SeaGreen3!10!} \cellcolor{white}  & A-BL & \(\mathbf{0.80}\) & \(\mathbf{35.3 \pm 24.9}\) & \(\mathbf{18.1 \pm 9.3}\) & \(\mathbf{1.2 \pm 0.5}\) \\ 
       \multirow{-3}{*}{\rotatebox[origin=c]{90}{Cuboid}}    & \textbf{ConPoSe} & \(\mathbf{0.80}\) & \(37.8 \pm 20.0\) & \(20.5 \pm 8.9\) & \(1.6 \pm 0.4\) \\ 
      \midrule    
      \rowcolor{SeaGreen3!10!} \cellcolor{white}  & N-LLM & \(\mathbf{0.87}\) & \(30.4 \pm 20.8\) & \(18.3 \pm 11.0\) & \(\mathbf{1.3 \pm 0.5}\) \\ 
         & A-BL & \(\mathbf{0.87}\) & \(\mathbf{27.3 \pm 16.9}\) & \(\mathbf{15.7 \pm 7.8}\) & \(1.4 \pm 0.3\)  \\ 
        \rowcolor{SeaGreen3!10!} \cellcolor{white} \multirow{-3}{*}{\rotatebox[origin=c]{90}{Cylinder}} & \textbf{ConPoSe} & \(0.80\) & \(28.4 \pm 17.7\) &  \(15.7 \pm 8.3\) & \(1.5 \pm 0.4\) \\ 
      \midrule 
        & N-LLM & \(0.20\) & \(22.0\pm 29.0\) & \(11.1 \pm 11.0\) & \(2.0 \pm 1.6\) \\ 
      \rowcolor{SeaGreen3!10!} \cellcolor{white}  & A-BL & \(\mathbf{0.47}\) & \(20.8 \pm 15.4\) & \(11.0 \pm 6.6\) & \(\mathbf{1.5 \pm 0.8}\) \\ 
      \multirow{-3}{*}{\rotatebox[origin=c]{90}{T-Shape}}  & \textbf{ConPoSe} & \(0.07\) & \(\mathbf{18.8 \pm 15.6}\) & \(\mathbf{8.2 \pm 5.4}\) & \(2.2 \pm 1.4\)  \\ 
      \bottomrule
     \end{tabular}
\end{table}

Finally, we evaluate the full approach shown in Fig.~\ref{fig:Overview}, including the contact point switching step (see Sec.~\ref{sec:cp_switching}).
Switching introduces additional error sources, as robots must operate in highly constrained spaces with other robots nearby, which may lead to insufficient space for re-positioning and a higher risk of inter-robot collisions. 
We test ConPoSe and the two baselines across all five scenarios and three objects with \(N=\{3, 5, 7\}\)~robots. 
Please note that all metrics in Tab.~\ref{tab:full_system} also include unsuccessful runs, possibly biasing execution time and the number of pushing configurations to be lower for scenarios with lower success rates. 

Success rates decrease for ConPoSe and the two baselines. 
All three methods have similar success rates for the cuboid and cylinder, with a maximum of one successful run difference. 
For the T-shape, however, we find huge drops in the success rates. 
This has probably two reasons: (i) the larger object size of the T-shape limits available free space for robot movement, and (ii) concave corners create bottlenecks where robots cannot switch contact points without an explicit avoidance strategy. 
We currently mitigate this issue by letting robots positioned at concave corners switch contact points consecutively. 
Despite these extra challenges, the analytical baseline still reaches success rates of \(60-80\,\%\) for \(N\in \{3, 5\}\), yet yields no successful runs for \(N = 7\). 
Both ConPoSe and the naive LLM-based baseline only result in one and three successful experiments, respectively.  
We attribute the advantage of the analytical baseline to two factors: 
First, it requires a lower number of pushing configurations~\(Z\), and second, it likely selects fewer pushing configurations with two or more robots closely positioned at a concave corner. 
Both aspects reduce the chances of failures during contact point switches. 
Since the performance for the T-shape was high for our experiments with automatic repositioning of robots (see Tab.~\ref{tab:shapes}), the decrease in performance can be attributed to the contact point switching method. 
Improving the switching algorithm remains an open challenge for future work. 

As expected, execution times increase across all scenarios. 
Switching itself is costly, taking \(1.2 - 2.2\,\text{min}\) per switch, which significantly contributes to the overall execution time and increases with the number of pushing configurations~\(Z\).
ConPoSe has lower~\(Z\) and thus also faster execution times than the naive LLM-based baseline. 
Still, it is slightly slower than the analytical baseline and requires a few more pushing configurations. 
As argued before, ConPoSe becomes competitive for \(N = 7\) and even outperforms the analytical baseline for \(N \ge 10\), see Fig.~\ref{fig:query-time-bar}. 
We are confident that the LLM-supported local search maintains its scalability advantage also when the switching mechanism is included.

\section{CONCLUSION}

In this work, we introduced ConPoSe, an LLM-guided local search method for contact point selection in a cooperative object pushing task. 
It consistently achieved short selection times, low numbers of pushing configurations, and high success rates across all tested MRS sizes. 
ConPoSe outperforms the analytical computation of contact points in time for \(N>7\), while maintaining competitive success rates. 
Despite limited prior knowledge about the object and considerable slippage, ConPoSe, as well as our two baselines, can successfully select suitable contact points for pushing different objects along a preplanned path.

However, the contact point switching mechanism is a key limitation. 
While approaches similar to ours have proven successful with holonomic robots~\cite{tang2024collaborativeplanarpushingpolytopic}, our round differential drive robots experience significant 
slippage, frequently losing their assigned contact points and restricting movement directions.
This complicates switching contact points and frequently results in inter-robot collisions. 
Addressing this challenge by developing a more robust switching strategy will be a central focus of our future work.

We are confident that our results can be replicated with physical robots based on preliminary experiments in real-world environments. 
Information that is currently provided by the simulation, including the object shape and its geometrical center, can be derived using modern AI methods, such as open-set object detection (e.g., Grounding Dino~\cite{liu2023grounding}) and segmentation (e.g., SAM~2~\cite{ravi2024sam2segmentimages}). 

\section*{ACKNOWLEDGMENTS}

The authors thank Lucas Kaiser for support in creating Fig.~\ref{fig:visualization_scenario}.


\end{document}